\pdfoutput=1

\documentclass[11pt]{article}

\usepackage[final]{acl}

\usepackage{times}
\usepackage{latexsym}
\usepackage{times}
\usepackage{latexsym}
\usepackage{algorithm}
\usepackage{algpseudocode}
\usepackage{amsmath}
\usepackage{hyperref}
\usepackage[T1]{fontenc}
\usepackage[utf8]{inputenc}
\usepackage{microtype}
\usepackage{enumitem}
\usepackage{inconsolata}
\usepackage{graphicx}
\usepackage{booktabs}

\usepackage[T1]{fontenc}

\usepackage[utf8]{inputenc}

\usepackage{microtype}

\usepackage{inconsolata}

\usepackage{graphicx}

\title{VeriMinder: Mitigating Analytical Vulnerabilities in NL2SQL}

\author{Shubham Mohole \\
Cornell University \\
\texttt{sam588@cornell.edu} \\\And
Sainyam Galhotra \\
Cornell University \\
\texttt{sg@cs.cornell.edu} \\}

\begin{document}
\maketitle

\begin{abstract}
Application systems using natural language interfaces to databases (NLIDBs) have democratized data analysis. This positive development has also brought forth an urgent challenge to help users who might use these systems without a background in statistical analysis to formulate bias-free analytical questions. Although significant research has focused on text-to-SQL generation accuracy, addressing cognitive biases in analytical questions remains underexplored. We present VeriMinder,\footnote{\href{https://veriminder.ai}{https://veriminder.ai}}, an interactive system for detecting and mitigating such analytical vulnerabilities. Our approach introduces three key innovations: (1) a contextual semantic mapping framework for biases relevant to specific analysis contexts (2) an analytical framework that operationalizes the Hard-to-Vary principle and guides users in systematic data analysis (3) an optimized LLM-powered system that generates high-quality, task-specific prompts using a structured process involving multiple candidates, critic feedback, and self-reflection.

User testing confirms the merits of our approach. In direct user experience evaluation, 82.5\% participants reported positively impacting the quality of the analysis. In comparative evaluation, VeriMinder scored significantly higher than alternative approaches, at least 20\% better when considered for metrics of the analysis's concreteness, comprehensiveness, and accuracy. Our system, implemented as a web application, is set to help users avoid "wrong question" vulnerability during data analysis. VeriMinder code base with prompts \footnote{\href{https://reproducibility.link/veriminder}https://reproducibility.link/veriminder} is available as an MIT-licensed open-source software to facilitate further research and adoption within the community. 
\end{abstract}

\section{Introduction}
\begin{figure}[t]  
    \centering       
    \includegraphics[width=\columnwidth]{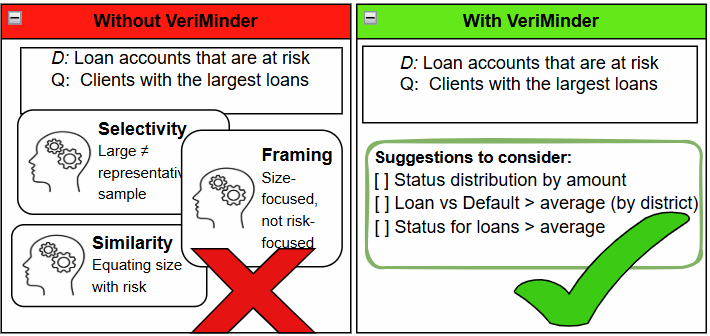}
    \caption{Example from experimental dataset showing VeriMinder mitigating biases via refinement suggestions}
    \label{fig:intro} 
\end{figure}
Natural Language to SQL (NL2SQL) systems have emerged as a critical technology for democratizing data access, enabling non-technical users to query complex databases without specialized SQL knowledge. However, this positive development is not without significant risks. A technically perfect SQL query derived from a fundamentally flawed analytical question will yield misleading results. Systems like SQLPalm ~\cite{sun2023sqlpalm}, SPLASH \cite{elgohary2020speak}, and DAIL-SQL \cite{gao2023texttosqlempoweredlargelanguage} focus on NL2SQL accuracy but do not consider the analytical quality of the user's original question.

Consider this example shown in Figure 1: A financial analyst tasked to identify ``loan accounts that are at risk'' but asks for ``clients with the largest loans.'' This query exhibits multiple cognitive biases: (1) \textit{Similarity bias} - incorrectly assuming that ``largest loans'' and ``at-risk loans'' are similar categories, (2) \textit{Framing bias} - framing the question around loan size rather than risk factors, completely changing what information will be retrieved, and (3) \textit{Selection bias} - focusing only on large loans selects a non-representative subset of potentially risky accounts, as small loans may have higher default rates. While a state-of-the-art NL2SQL system can generate syntactically correct SQL for the original question, it cannot address these analytical blindspots, leaving a critical vulnerability unaddressed.  

Research shows cognitive biases significantly impact professional decision-making across fields like medicine and laws ~\cite{berthet2022impact}. The consistent association of these biases, such as anchoring and availability, with detrimental outcomes like health diagnostic inaccuracies underscores the critical need for mitigation systems like VeriMinder. As Peter Drucker said, ``The most serious mistakes are not being made due to wrong answers. The truly dangerous thing is asking the wrong question.''~\cite{drucker1971men}.

Traditional approaches to mitigating such issues rely on static checklists~\cite{lenders2025users} or educational interventions ~\cite{thompson2023educational}, which are challenging to implement consistently. While FISQL \cite{menon2025fisql} and SPLASH \cite{elgohary2020speak} offer limited feedback mechanisms, they focus primarily on SQL refinement rather than addressing analytical quality issues ~\cite{qu2024generationalignitnovel}.
 
 To address these challenges, we present VeriMinder, which identifies and mitigates analytical vulnerabilities in NL2SQL workflows. Our interactive web application addresses these vulnerabilities with three innovations: (1) a semantic framework that systematically detects biases and blindspots in analytical questions; (2) a structured analytical process based on the "Hard-to-Vary" principle~\cite{deutsch2011beginning}; and (3) an optimized LLM-driven refinement interface, integrated with NL2SQL workflows. VeriMinder integrates seamlessly with existing NL2SQL systems through simple configuration, supporting users of such systems with robust analytical question formulation alongside accurate SQL generation. Our evaluation demonstrates that VeriMinder significantly enhances analytical outcomes, outperforming baseline approaches across key analytical metrics. 
\section{System Architecture}
VeriMinder operationalizes Deutsch's \textbf{Hard-to-Vary principle}~\cite{deutsch2011beginning} through a systematic architecture to identify and mitigate analytical vulnerabilities in user questions ($Q$), transforming potentially biased queries into robust analytical explanations ($E$) within a given domain ($D$) and decision context ($C$). This principle posits that good explanations are constrained, such that altering their components weakens the explanations or creates inconsistency. Applied to data analytics, a robust explanation $E$, often operationalized via SQL queries ($S$), is hard-to-vary if its components necessarily and cohesively address $Q$ in context $C$, lacking arbitrary elements whose removal wouldn't degrade quality. Easily varied explanations, conversely, allow interchangeable components without specific roles, potentially leading to misleading results from flawed questions (e.g., analyzing broad expense categories instead of particular cost drivers while deciding on governmental cost-cutting measures). VeriMinder enforces this by ensuring the analysis pinpoints specific factors, yielding data-supported, falsifiable explanations that resist variation.

\subsection{Core Modules and Architecture}

\begin{figure}[ht!]  
    \centering       
    \includegraphics[width=\columnwidth]{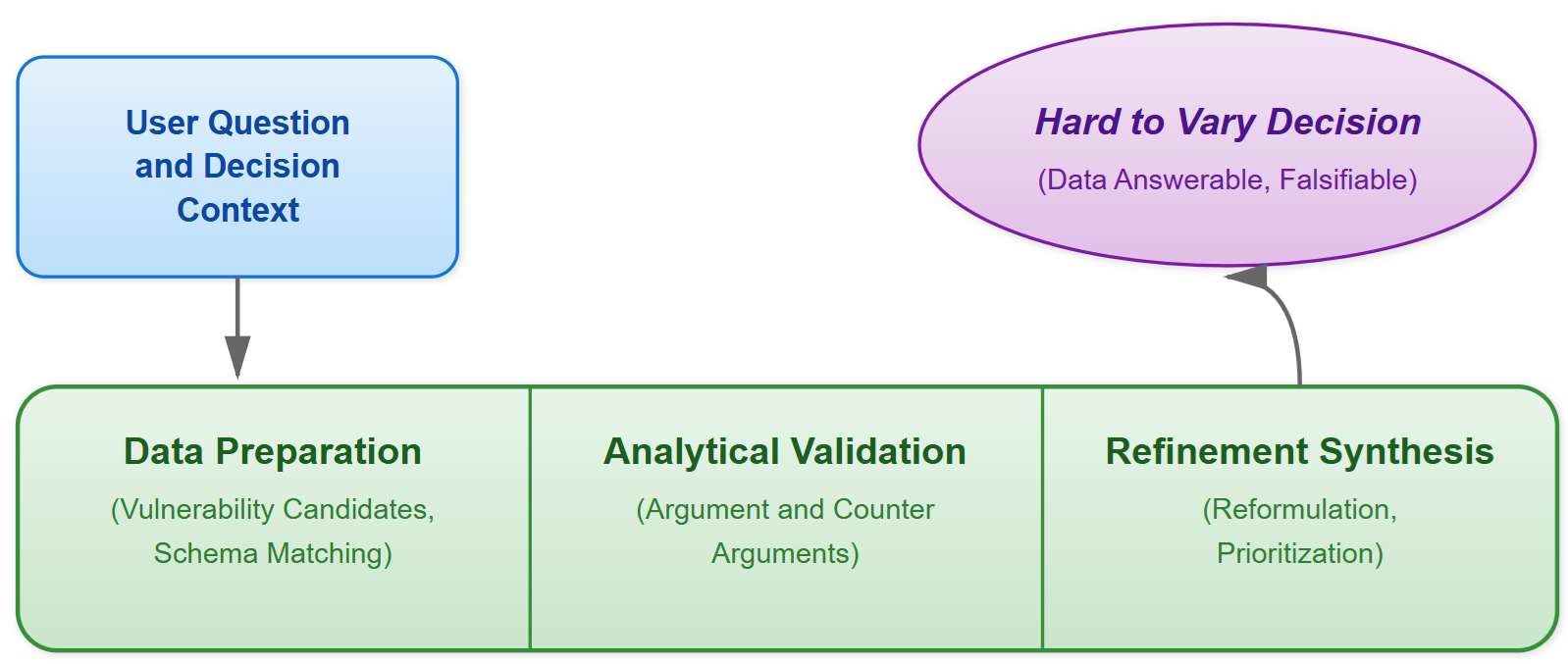}
    \caption{Three-stage framework operationalizing the Hard-to-Vary principle.}
    \label{fig:framework} 
\end{figure}

The VeriMinder system implements a systematic approach that helps analysts refine vulnerable questions into robust data analysis to operationalize the hard-to-vary principle. As shown in Figure 2, our architecture processes natural language questions through three sequential stages: Data Preparation, Analytical Validation, and Refinement Synthesis.

The system analyzes the question and decision context in the data preparation stage to identify potential analytical vulnerabilities and relevant schema elements. During Analytical Validation, vulnerabilities are detected, and structural analysis is performed using argument components and counter-argument testing to verify their significance. In Refinement Synthesis, the system generates targeted refinement suggestions that help with analysis aligned with a hard-to-vary approach for data-backed explanations for the particular decision context.

\begin{figure}[t]  
    \centering       
    \includegraphics[width=\columnwidth]{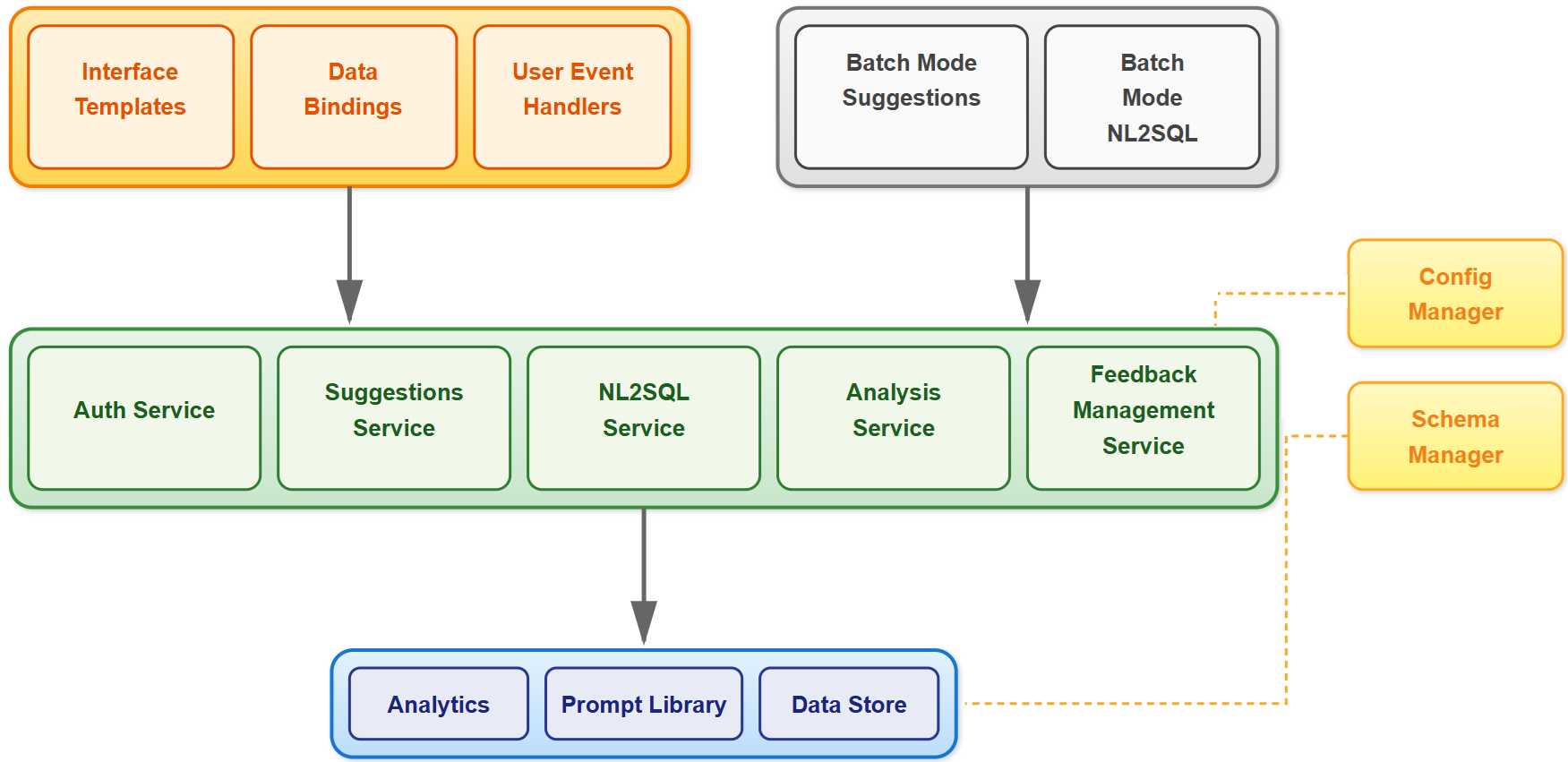}
    \caption{Modular architecture supporting scalability and flexible deployment modes}
    \label{fig:system_modules} 
\end{figure}

VeriMinder implements this framework using a modular service-based architecture (Figure 3) for flexibility, featuring five core services communicating via standardized interfaces: \textbf{Auth} (user provisioning/access, future enterprise plugins), \textbf{Suggestion} (implements core framework analytics), \textbf{NL2SQL} (extends the approach from ~\cite{qu2024generationalignitnovel} with metadata and dataset-specific distribution information and uses Gemini Flash 2.0 ~\cite{gemini2025}), \textbf{Analysis} (compares initial vs. refined results for user reflection), and \textbf{User Feedback} (collects improvement data). The underlying analytical framework components (detailed in Appendix A.1) comprise 53 categorized cognitive biases (e.g., Memory, Statistical, Framing), data schema patterns (temporal, categorical, numerical detailed in Appendix A.2), the Toulmin model for argument structure evaluation ~\cite{toulmin1958uses} (Appendix A.3), and counter-argument frameworks ~\cite{greitemeyer2023counter} for questions that help address challenges and refine explanations (Appendix A.4).

For our system implementation, we developed an experimental NL2SQL component based on best practices for LLM-based text-to-SQL generation~\cite{qu2024generationalignitnovel, sun2023sqlpalm, gao2023texttosqlempoweredlargelanguage}. VeriMinder is designed to complement existing NL2SQL systems rather than replace them, focusing on the orthogonal problem of analytical question formulation.

\subsection{Prompt Formulation Method}
\begin{figure}[ht]
    \centering
    \includegraphics[width=\columnwidth]{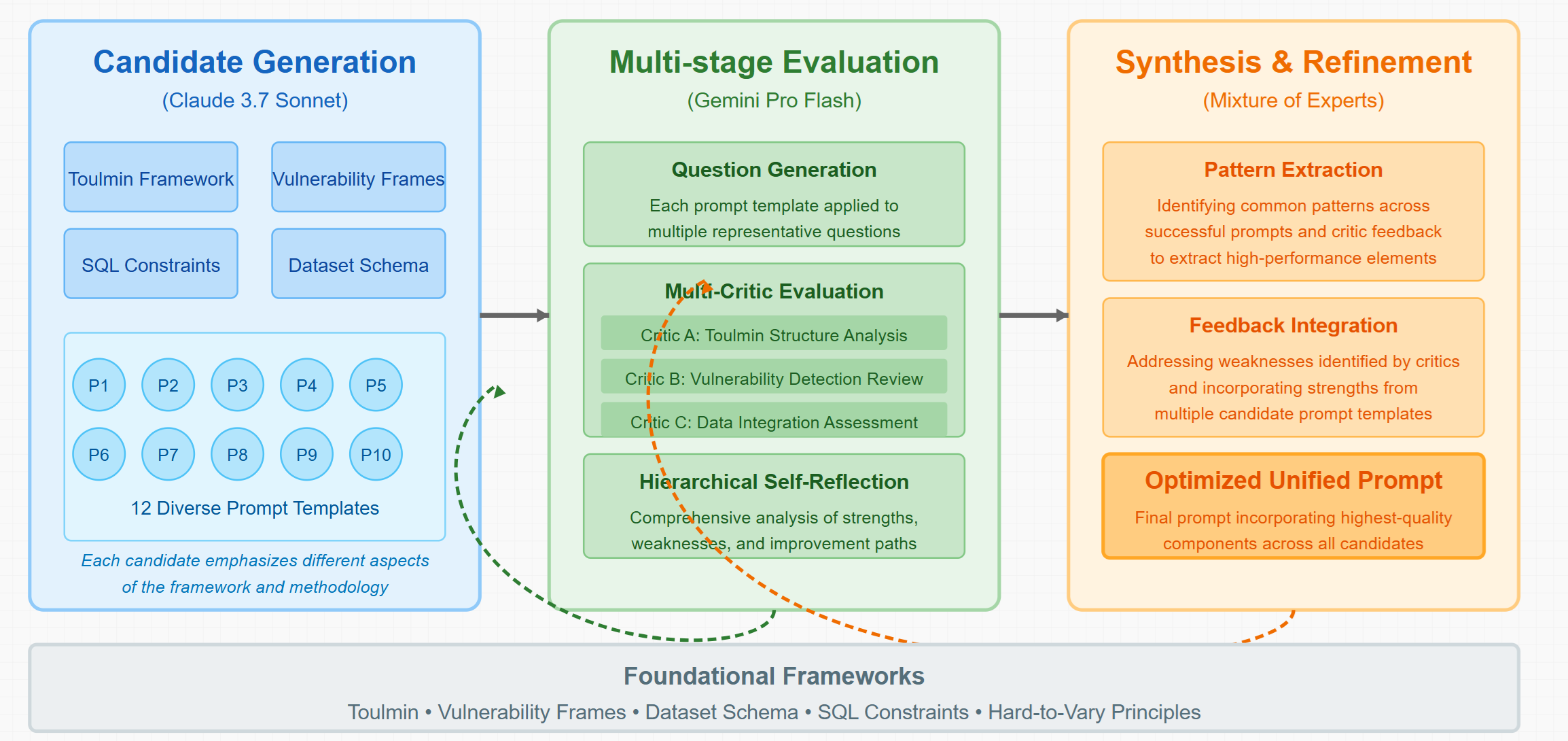}
    \caption{Multi-candidate prompt engineering pipeline with critic feedback and self-reflection.}
    \label{fig:prompt_design}
\end{figure}

VeriMinder offers users for their free-form analytical questions bias-mitigating alternatives through a three-stage workflow (Figure~\ref{fig:prompt_design}). The pipeline is driven by a formally defined hard-to-vary objective but is implemented with practical approximations that respect LLM limits and inference latency.

\subsubsection{Information-Theoretic Grounding}
\label{sec:framework}
The architecture of VeriMinder is guided by a core principle: a robust analytical question should maximize predictive insight about a decision while minimizing its own descriptive complexity, subject to an interactive-latency budget. This section outlines the ideal theoretical framework that motivates our system's design (\S\ref{ssec:ideal}) and details its translation into a practical, multi-stage LLM pipeline (\S\ref{ssec:proxies}-\ref{ssec:refinement}), concluding with a discussion of its scope and limitations (\S\ref{ssec:limitations}).

\subsubsection{Idealized Theoretical Motivation}
\label{ssec:ideal}
We formalize the principle of robust inquiry using the Hard-to-Vary (HV) score, a metric inspired by Deutsch's concept of good explanations~\cite{deutsch2011beginning} and the Minimum Description Length (MDL) principle~\cite{rissanen1978modeling, grunwald2007minimum}. For a set of selected analytical variables, $S$, and a decision target, $T$, the HV score is:
\begin{equation}
\label{eq:hv_score}
HV(S) = \frac{I(T;S)}{DL(S)}
\end{equation}
Here, $I(T;S)$ is mutual information~\cite{cover2006elements}, and $DL(S)$ is the model's description length. This formulation, which extends normalized information metrics like the Information Gain Ratio~\cite{quinlan1993c4}, rewards explanatory density (high information per unit of complexity) and echoes the objective of Information Bottleneck theory~\cite{tishby2000information}.

To verify this metric's behavior, we developed a numeric validation suite. As detailed in our code repository, experiments on synthetic Bayesian networks demonstrate the HV score's key properties under idealized conditions. All simulations use an exact mutual information computation and define complexity as the variable set cardinality, i.e., $DL(S) = |S|$. This provides empirical support that the HV score is a sound theoretical target.

\subsubsection{Practical Heuristic Proxies}
\label{ssec:proxies}
Directly optimizing Eq. \ref{eq:hv_score} is computationally intractable even in structured feature spaces \cite{Vinh2014Effective}, and becomes exponentially more complex in the open-ended natural language domain where the search space includes all possible question formulations. VeriMinder therefore employs LLM-based \textit{heuristic proxies} guided by the HV formula's intuition. We recognize this is not a formal equivalence; the desirable properties of the HV score hold exactly only under the formal definition, while our proxies aim to approximate them empirically.
\begin{itemize}
    \item \textbf{LLM Critic Scores for $I(T;S)$:} We use scores from specialized LLM critics as a proxy for information value. The rationale is that high-quality questions (judged on insight, logic, and bias mitigation) are more likely to reduce uncertainty about the decision target. This aligns with Information Foraging Theory~\cite{pirolli1999information} and the use of LLMs as evaluators~\cite{zheng2023judging, dubois2023alpacafarm}.
    \item Motivated by evidence that excessive prompt length can degrade LLM reasoning~\cite{jiang2023longllmlingua}, our prompt templates are built around a concise, analytical flow that goes from context analysis to final question selection, designed to produce a minimal set of high-impact questions. We therefore model task complexity through this structured analytical process rather than raw token count.
\end{itemize}

\subsubsection{Stage 1: Ensemble-based Candidate Generation}
To explore the analytical space, the system using generates a diverse set of candidates using twelve prompt templates. These templates are themselves the output of an automated meta-level prompt engineering process based on Claude 3.7 Sonnet model \cite{anthropic2025claude37} selected for its intelligence category rank \cite{artificialanalysis2025}), ensuring each targets a distinct analytical angle (e.g., vulnerability detection, schema validation). This ensemble method ensures broad coverage, a technique well-grounded in machine learning for both bagging~\cite{breiman1996bagging} and modern LLM prompting~\cite{zhou2023largelanguagemodelshumanlevel}.

\subsubsection{Stage 2: Distributed Critic Evaluation}
Generated candidates are evaluated by a panel of three specialized LLM critics (based on the Claude 3.7 Sonnet model). For efficiency, a random subset of \textbf{two} critics evaluates each candidate. This implements distributed evaluation analogous to boosting, where a committee of weak learners forms a robust judgment~\cite{schapire1990strength}. This aligns with modern methods using self-consistency and multi-agent consensus to improve LLM evaluation~\cite{wang2022self, li2024improving}.

\subsubsection{Stage 3: Critic Feedback and Self Reflection}
\label{ssec:refinement}
Finally, the system performs a single self-reflection pass that improves prompts using critic feedback.  This mirrors self-refinement techniques that improve LLM performance~\cite{madaan2023self, shinn2023reflexion}. At present we execute only one iteration but multiple self-reflection rounds would be a possible natural extension to the current pipeline.  

\subsubsection{Scope and Limitations}
\label{ssec:limitations}
Our approach has three main limitations. First, our production system relies on heuristic search, unlike the exhaustive search in our validation suite. Second, critic scores and our analytical flow stages are pragmatic surrogates, not formal equivalents, for $I(T;S)$ and $DL(S)$. Finally, our current cost model is limited to response structure and does not yet incorporate computational latency.

\subsection{Interactive User Interface}
\begin{figure}[ht!]
\centering
\fbox{\includegraphics[width=\columnwidth]{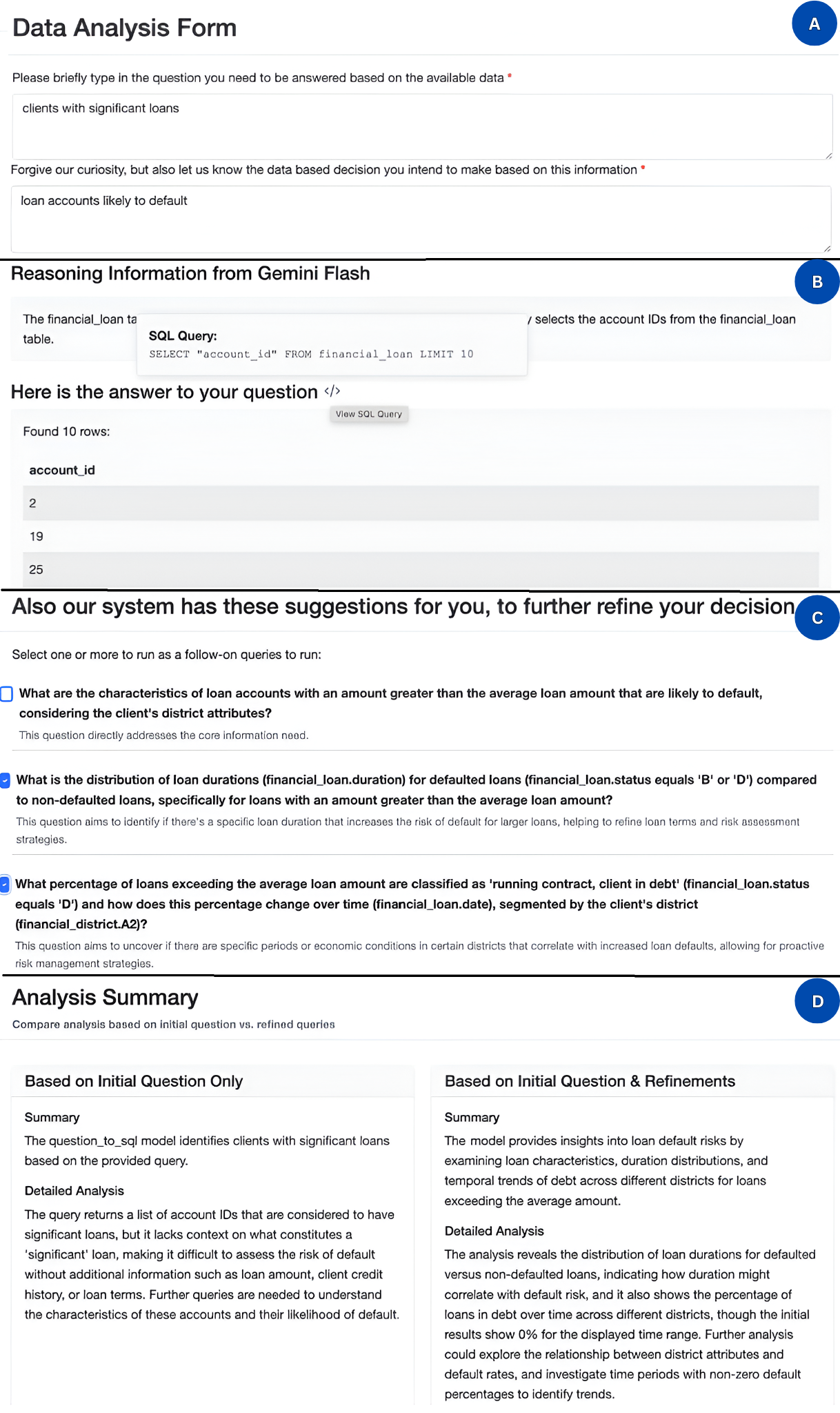}}
\caption{VeriMinder's user interface workflow: (A) Initial Question, (B) Query Results, (C) Refinement Suggestions, (D) Comparative Analysis}
\label{fig:interface}
\end{figure}

VeriMinder's user interface (Figure 5) employs a progressive disclosure pattern for a guided workflow: users provide their questions and context, the system executes the query while analyzing vulnerabilities, suggests refinements for user selection, presents a side-by-side comparison of results, and explains detected issues and fixes. To enhance user experience during intensive computations, server-sent events (SSE) provide streaming updates and educational insights. The system features a pluggable interface and unified abstraction layer to support multiple database types, utilizing SQLite (with the BIRD-DEV benchmark \cite{li2023llm}) for execution and MySQL for tracking application state.

\section{Experiments}
\subsection{Experimental Setup}
To comprehensively evaluate VeriMinder, we designed a multi-step assessment framework addressing key research questions: (1) How effective is the VeriMinder solution in improving the analysis using the NL2SQL interface? (2) How does our approach compare with alternative methods for enhancing analytical quality on key accuracy, concreteness, and comprehensiveness metrics ~\cite{zhu2024situatednaturallanguageexplanations}?

The evaluation dataset was derived from the BIRD-DEV benchmark questions. To create realistic decision contexts, we manually crafted the 164 decision scenarios following the Case Study Method~\cite{ellet2007case}, ensuring balanced coverage of choice, evaluation, and diagnosis types. Data analytics experts designed these scenarios to represent contexts where analytical vulnerabilities could significantly impact outcomes. We employed TF-IDF vectorization to match each decision with the most semantically relevant question from BIRD-DEV, creating a bipartite relationship. The final decision text was lightly edited for grammar and sentence structure to ensure consistency during the user study without altering the analytical focus of the decision contexts.  This methodical approach yielded 164 question-decision pairs, divided into three subsets: 64 pairs (DS1) for human evaluations and 100 pairs (DS2) for automated assessment. An additional smaller subset DS1-T1 of 36 pairs was created from the DS1. All splits were done randomly.

To our knowledge, no direct comparable system focuses on refining user-posed questions and addressing biases and blind spots. So in addition to the \textbf{Direct NL2SQL} (standard text-to-SQL generation without analytical enhancements), we evaluated VeriMinder by operationalizing three alternative approaches that the research community has considered for either bias mitigation or holistic analysis: \textbf{Decision-Focused Query Generation} (generating questions directly from decision context~\cite{zhang2025llmsdesigngoodquestions}), \textbf{Question Perturbation (PerQS)} (creating variations of the original question~\cite{zhu}), and \textbf{Critic-Agent Feedback (CAF)} (implementing a critic agent providing feedback~\cite{li2024critical}).  We use the same LLM (Gemini Flash 2.0) for all baselines as VeriMinder and plan to release them as part of our code release.

A critical aspect of our evaluation methodology was ensuring consistent SQL generation across all compared systems. To isolate the effect of analytical question formulation (our focus) on NL2SQL accuracy, we implemented the same experimental NL2SQL component for all baseline systems and VeriMinder. For our evaluations, we validated that all generated SQL queries executed correctly before assessment, allowing us to focus purely on analytical quality rather than technical SQL correctness.

\subsection{User Experience Evaluation}
We conducted an interactive user study with the DS1-T1 dataset, recruiting 63 participants from Prolific ~\cite{prolific2025} with diverse backgrounds. For 30 scenarios, we received submissions from two users each, and for three scenarios, from one user (a total of 63 unique participants). Appendix B1 shows the feedback form presented to participants. The overall effectiveness of our solution in improving analysis quality received 82.5\% positive ratings (score of 4 or 5), with Gwet's AC1 of 0.766. Suggestion effectiveness received 74.6\% positive ratings, with Gwet's AC1 0.670. Rationale clarity had 66.7\% (Gwet's AC1 0.479) and Scenario realism 61.9\% (Gwet's AC1 0.457) positive ratings. The reliability scores, particularly for clarity and realism, likely reflect the diverse user base from Prolific. Furthermore, the scenario realism scores may be influenced by the experimental setup, where decision contexts were constrained by matching them to the existing BIRD-DEV dataset questions.

\subsection{Comparative System Evaluation}
From the DS1 dataset, we conducted a comparative evaluation of generated analysis questions with one data analyst from each of the two US-based software companies who responded to our request. Appendix B.2 shows the screenshot of the interface these data analyst users used to rate the comparative strength of analysis questions in a decision context. As with the previous test, we only included the successful completions in our analysis (because of an unrelated system outage issue, we failed to get submissions for five entries). For the 59 scenarios, we received submissions from both users. VeriMinder demonstrated strong performance across all dimensions: Accuracy (mean=7.87/10, 95\% CI [7.57, 8.18]), Concreteness (mean=7.79/10, 95\% CI [7.47, 8.10]), and Comprehensiveness (mean=8.05/10, 95\% CI [7.74, 8.36]).

\begin{figure}[t]
\centering
\includegraphics[width=\columnwidth]{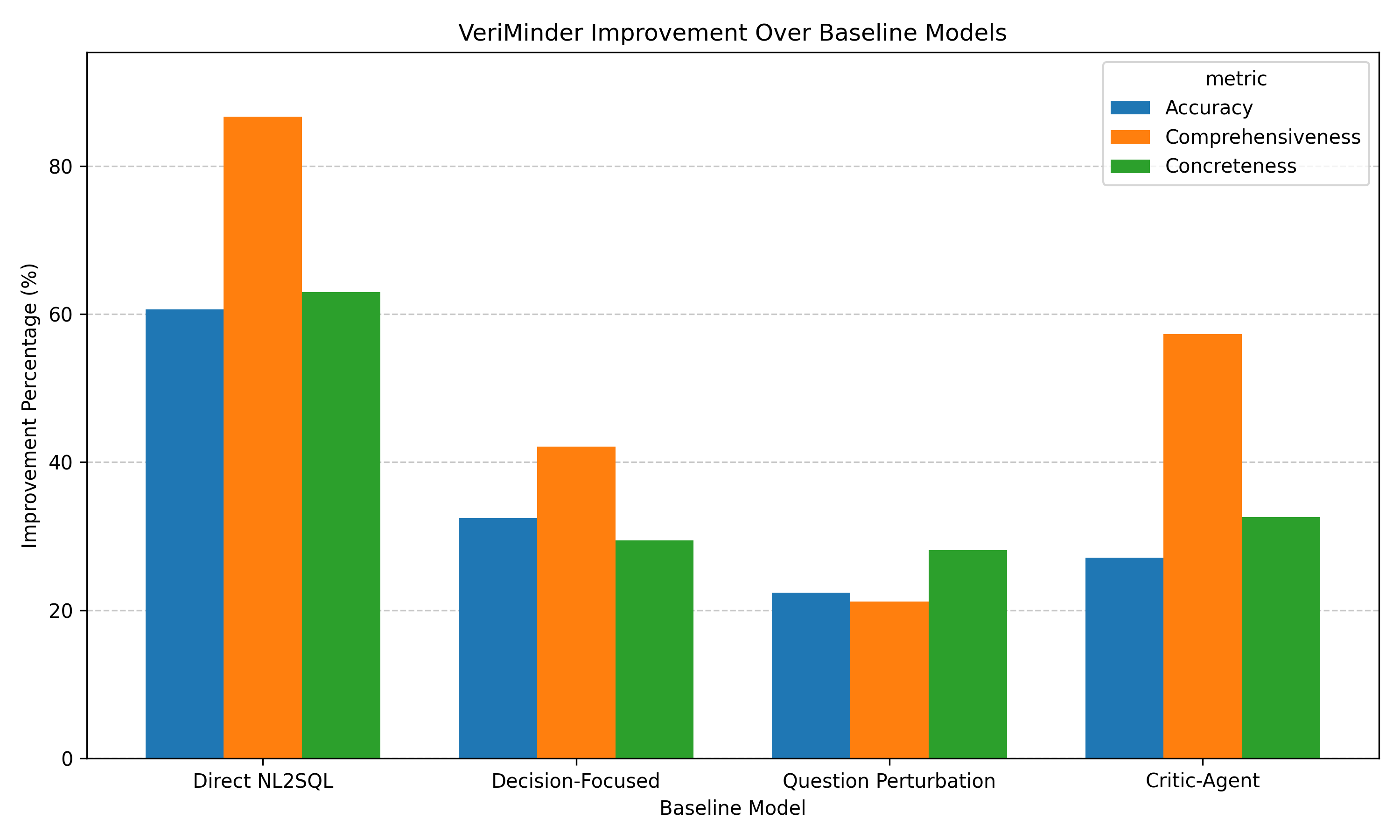}
\caption{Percentage improvement of VeriMinder over baseline systems on key analytical dimensions}
\label{fig:improvement}
\end{figure}

Figure~\ref{fig:improvement} illustrates VeriMinder's percentage improvement over each baseline system. The most substantial improvements were observed against Direct NL2SQL, with gains of 60.4\% in Accuracy, 63.2\% in Concreteness, and 86.9\% in Comprehensiveness. Even against the strongest baseline (Question Perturbation), VeriMinder showed improvements of 22.1\% in Accuracy, 28.4\% in Concreteness, and 21.2\% in Comprehensiveness.

Statistical analysis confirmed these improvements were significant (p < 0.001) with paired t\_test across all dimensions and baseline comparisons. Win rates further illustrated VeriMinder's quality, outperforming Direct NL2SQL in 83.9\% of Accuracy comparisons, 86.4\% of Concreteness comparisons, and 97.5\% of Comprehensiveness comparisons. Inter-rater reliability metrics based on the model ranks demonstrated robust agreement in our evaluations, with Gwet's AC1 coefficients of 0.941 for Accuracy, 0.960 for Concreteness, and 0.862 for Comprehensiveness.

\subsection{Large-Scale Automated Evaluation}
We employed an LLM-based evaluator for dataset DS2 (100 scenarios) (Gemini Flash 2.0).  With known limitations of LLM for quantitative scoring ~\cite{openai2024gpt4technicalreport, bubeck2023sparksartificialgeneralintelligence} but better performance in verbal analysis and relative ranking \cite{zheng2023judging, Gilardi_2023}, our test focused on LLM skills in text comprehension and comparative qualitative assessments. In Appendix B.3, we discuss our approach to the prompt design. For LLM-based evaluation, we first calibrated our automated evaluator (based on Gemini 2.0 Flash) against human judgments on comparative ranking on a subset of 15 examples from DS1, finding a m (Pearson's $r=0.74, p<0.001$) that provided us confidence in the automated results.

\begin{figure}
\centering
\includegraphics[width=\columnwidth]{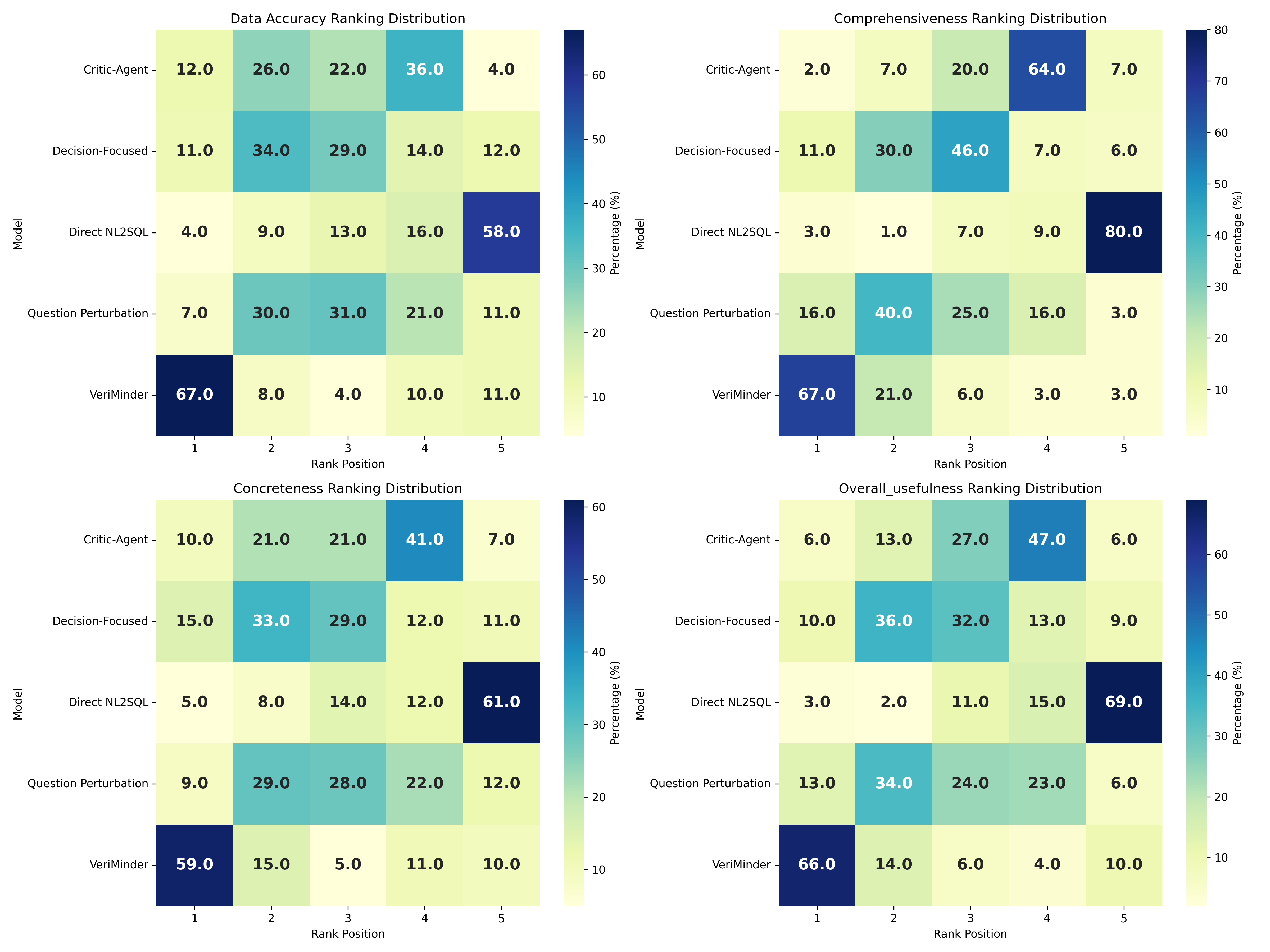}
\caption{Ranking distribution across analytical dimensions; VeriMinder consistently achieves highest rankings}
\label{fig:ranking_distribution}
\end{figure}

As Figure 7 shows, VeriMinder consistently achieved the highest first-place rankings: 67.0\% for Data Accuracy, 67.0\% for Comprehensiveness, 59.0\% for Concreteness, and 66.0\% for Overall Usefulness. In contrast, Direct NL2SQL received the most last-place rankings across all metrics, highlighting the importance of analytical enhancement beyond raw SQL generation.

\subsection{Analysis of Bias Mitigation Effectiveness}

\begin{figure}[ht!]
\centering
\includegraphics[width=\columnwidth]{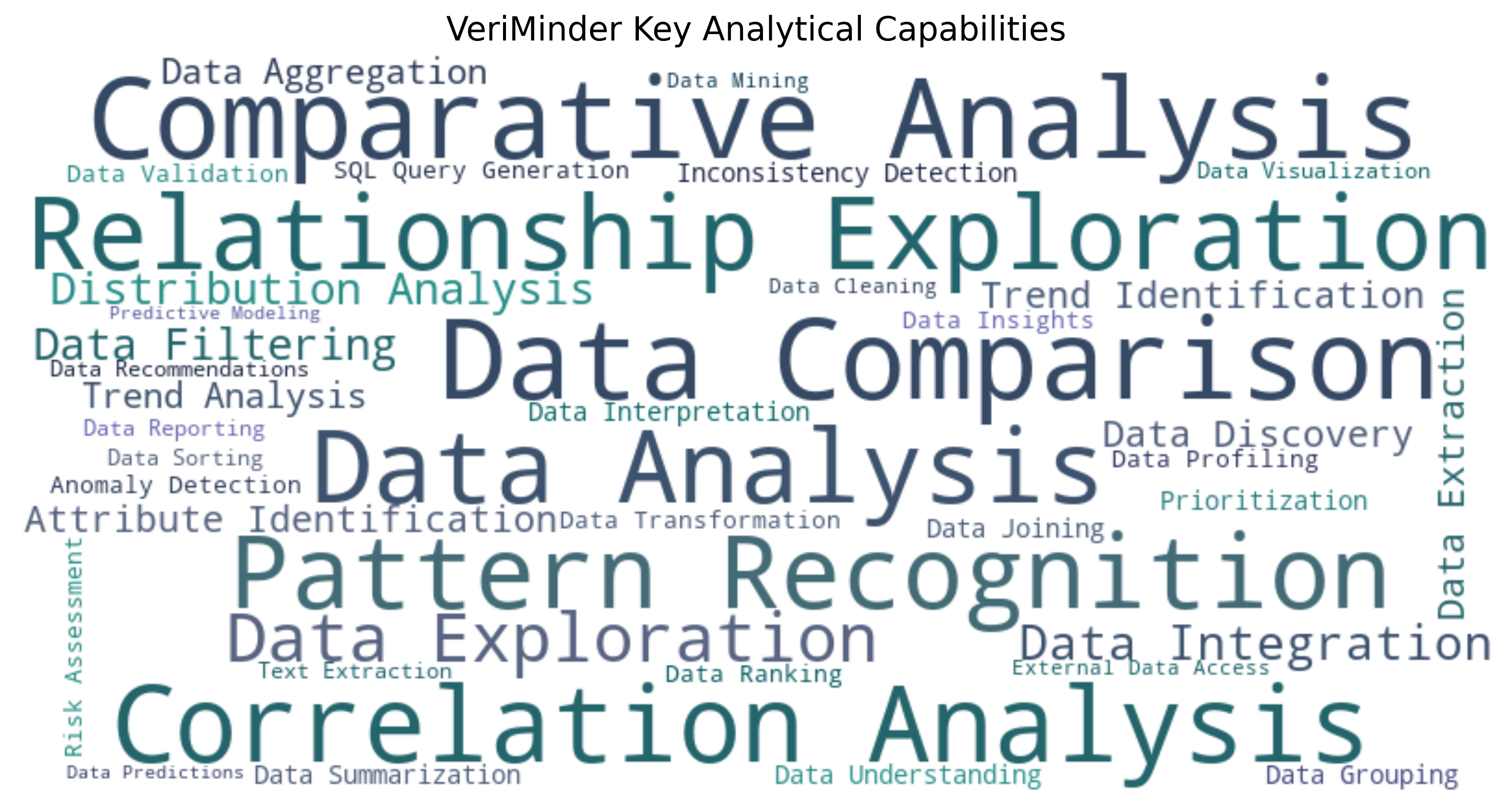}
\caption{Key analytical capabilities driving cognitive bias mitigation in VeriMinder}
\label{fig:wordcloud}
\end{figure}

The word cloud visualization in Figure \ref{fig:wordcloud} highlights VeriMinder's key analytical capabilities as identified through qualitative analysis of LLM response. This visualization was generated through automated content analysis of refinement suggestions across the dataset. As shown in Figure 8, comparative analysis, pattern recognition, and relationship exploration emerge as key capabilities, enabling VeriMinder to mitigate cognitive biases.

\subsection{Limitations}
Several limitations should be noted. First, deployment in specific domains may require customization of the analytical components. Second, the system's effectiveness depends on the underlying NL2SQL engine quality, implemented here as a simplified service module. We evaluated VeriMinder primarily on BIRD-DEV, which LLMs may have seen during training, raising concerns about information leakage and overestimated SQL success rates on truly unseen databases. The interface is desktop-optimized without accessibility testing. Before general release, critical enhancements include mobile support, accessibility features, multi-query handling, and validation on previously unseen databases to confirm generalization capabilities.

\section{Related Work}
Our work builds upon research across cognitive bias mitigation, natural language database interfaces, and LLM reasoning techniques in non-ground truth regimes - analytical contexts where there is no single 'correct' answer but varying degrees of analytical quality based on comprehensiveness, accuracy and alignment with decision objectives. Prior work in cognitive bias mitigation has examined biases in data-driven contexts ~\cite{kahneman2011thinking, tversky1974judgment, sumita2024cognitivebiaseslargelanguage, ke2024mitigating}, but primarily focused on bias awareness rather than active mitigation within analytical workflows. Benchmarks like Spider 2 \cite{lei2025spider20evaluatinglanguage} have driven recent advancements in NL2SQL generation ~\cite{deng2025reforcetexttosqlagentselfrefinement, wang2025linkalignscalableschemalinking}, with LLM-based systems achieving high execution accuracy. However, these systems primarily address technical SQL issues rather than analytical vulnerabilities. 

While VeriMinder primarily focuses on analytical question formulation, our evaluation employs a simplified NL2SQL service. This service incorporates metadata and dataset-specific distribution information for  SQL generation within our setup, drawing inspiration from recent work on mitigating NL2SQL hallucinations, such as the Task Alignment strategy proposed by~\cite{qu2024generationalignitnovel} and LLM based tabular learning tasks enhanced through \cite{mohole2025sifotl} columnar statistics for datasets. LLM prompting techniques, including response selection ~\cite{zhao2025samplescrutinizescaleeffective}, have enhanced reasoning capabilities but might not be suitable for a non-ground truth regime that requires an interactive experience. With our principled approach, inspired by Deutsch's framework~\cite{deutsch2011beginning}, and a multi-candidate refinement process, we provide a lightweight yet systematic framework for optimizing LLM response for downstream NL2SQL and analysis tasks.

\vspace{-0.2em}

\section{Future Work and Conclusion}

While VeriMinder currently targets NL2SQL interactions, its analytical core is modality-agnostic, enabling future extensions to Python/pandas code generation for statistical exploration. Building on Self-RAG~\cite{asai2023selfraglearningretrievegenerate}, we plan to evolve our self-reflection phase into a multi-head, bias-aware rubric outputting calibrated probabilities for evidence sufficiency, cognitive-bias flags, and statistical validity. These probabilities will both steer an adaptive retriever-generator loop and serve as bias-aware non-conformity scores for Conformal LM~\cite{quach2024conformallanguagemodeling}, enabling rejection thresholds that preserve coverage while reducing bias. Our Information-Theoretic Framework extends naturally to this calibration focus—by maximizing $\text{HV}(S)$ over reflection head outputs, Information theory guided pruning could guarantee minimal causal sufficiency while keeping calibration lean.

With VeriMinder, we've presented an end-to-end system for  mitigating analytical vulnerabilities in NL queries. By operationalizing the "hard-to-vary" explanations we demonstrated its effectiveness for the NL2SQL use cases. Coupled with SELF-RAG principles and bias-aware Conformal prediction, this research can open avenues for NLIDBs that provide answers not only \emph{probably correct} but also \emph{unbiased and grounded in evidence}.

\section{Broader Impact Statement}

\textit{While VeriMinder addresses analytical vulnerabilities, key limitations, and ethical points remain:}

\paragraph{Analytical Guidance vs. Guarantee} The system offers guidance, not guarantees, enhancing but not replacing user critical thinking. Vulnerability detection may not be exhaustive.

\paragraph{Commercial API Dependencies} Reliance on commercial LLMs limits accessibility; future work should explore open-source alternatives.

\paragraph{Cultural and Domain Biases} The bias taxonomy is primarily Western-based and may need domain-specific or cultural adaptation.

\paragraph{Potential for Misuse} Analytical enhancement tools could be misused; governance frameworks are needed to ensure integrity.

\paragraph{Augmentation vs. Automation} VeriMinder augments human analysis, preserving user agency rather than fully automating the process.

\textit{We believe addressing analytical vulnerabilities is vital as data access is democratized. VeriMinder is an initial step aiming to inspire further research at the intersection of cognitive science, data analytics, and NLP.}

\bibliography{custom}

\newpage
\appendix
\renewcommand{\thesection}{\appendixname~\Alph{section}}
\renewcommand{\thesubsection}{\Alph{section}.\arabic{subsection}}


\section{Analytical Framework Components}
\label{sec:analytical-framework}

Our framework integrates four complementary analytical perspectives via an optimized LLM prompt to identify and mitigate vulnerabilities (biases, data mismatches, logical flaws, framing issues) in natural language queries before SQL generation.

\vspace{-4pt} 
\subsection{Cognitive Biases Framework}
\label{subsec:cognitive-biases}
\vspace{-4pt} 
Incorporates 53 cognitive biases relevant to data analysis ~\cite{soprano2024cognitive, dimara2020task, hilbert2012toward, caverni1990cognitive, ehrlinger2016decision}, mapping NL query patterns to potential reasoning pitfalls. Categories include:

\noindent\textbf{1. Memory Biases (8):} Hindsight, Imaginability, Recall, Search, Similarity, Testimony, False Memory, Availability.

\noindent\textbf{2. Statistical Biases (9):} Base Rate Neglect, Chance, Conjunction, Correlation, Disjunction, Sample Size Neglect, Subset Bias, Gambler's Fallacy, Probability Neglect.

\noindent\textbf{3. Confidence Biases (8):} Completeness Illusion, Illusion of Control, Confirmation Bias, Desire Bias, Overconfidence, Redundancy Illusion, Dunning-Kruger Effect, Bias Blind Spot.

\noindent\textbf{4. Methodological Biases (12):} Data Quality Neglect, Multiple Testing Fallacy, Selection Bias, Method Fixation, Tool Overconfidence, Selectivity, Success/Self-Serving Bias, Test Inability, Anchoring, Conservatism, Reference Dependence, Regression to Mean.

\noindent\textbf{5. Framing \& Contextual Biases (16):} Framing Effect, Linear Assumption, Mode Influence, Order Effect, Scale Distortion, Primacy Effect, Recency Effect, Granularity Illusion, Attenuation Bias, Complexity Avoidance, Escalation of Commitment, Habit, Inconsistency, Rule Adherence, Fundamental Attribution Error, Bandwagon Effect.

\vspace{-4pt}
\subsection{Data Schema Patterns}
\label{subsec:data-schema}
\vspace{-4pt}
Examines NL query alignment with data types. Key NL2SQL considerations:
\textbf{Temporal:} Handling date/time formats (e.g., `DATEPART`), consistent aggregation. 
\begin{enumerate}[itemsep=-2pt]
    \item \textbf{Categorical:} Resolving ambiguity (e.g., `LA` vs `Los Angeles`), implicit hierarchies. 
    \item \textbf{Numerical:} Interpreting average/median correctly (e.g., `AVG`), handling outliers. 
    \item \textbf{Relationship:} Inferring `JOIN` paths, verifying functional dependencies (e.g., city $\rightarrow$ zip). 
    \item \textbf{Data Quality:} Assessing missing data (`NULL`, `COALESCE`), inconsistencies (e.g., negative counts). 
    \item \textbf{Transformation:} Needs for normalization (per capita), discretization (`CASE WHEN`), aggregation (`GROUP BY`).
\end{enumerate}

\vspace{-4pt}
\subsection{Toulmin Argument Structure}
\label{subsec:toulmin}
\vspace{-4pt}
Evaluates the implicit argument in the NL query/SQL based on Toulmin's model~\cite{toulmin1958uses}:
\begin{enumerate}[itemsep=-2pt]
    \item \textbf{Claim Clarity/Relevance:} Does SQL capture NL assertion and align with context? (`SELECT`, `WHERE`). 
    \item \textbf{Evidence Sufficiency/Validity:} Enough reliable data retrieved? (`COUNT`, `LEFT JOIN`). Trustworthy sources?
    \item \textbf{Warrant Validity/Applicability:} Is NL-to-SQL logic sound? Respects constraints? (CTEs, domain checks).
    \item \textbf{Backing:} Logic supported by standard practices/definitions?. 
    \item \textbf{Qualifier Precision/Scope:} Acknowledges limits (confidence, scope `WHERE`, rounding)?. 
    \item \textbf{Rebuttal Considerations:} Alternative queries, interpretations (`JOIN` confounders), exceptions (`EXCLUDE`)?.
\end{enumerate}
\vspace{-4pt}
\subsection{Counter-Argument Frameworks}
\label{subsec:counter-argument}
\vspace{-4pt}
Systematically challenges the NL query/formulation for analytical rigor:
\begin{enumerate}[itemsep=-2pt]
    \item \textbf{Conclusion Rebutters:} Scope limitation needed? Alternative queries yield different conclusions? 
    \item \textbf{Premise Rebutters:} Relies on inaccurate/incomplete (`IS NULL`)/non-representative data? Metric appropriate?
    \item \textbf{Argument Undercutters:} Hidden assumptions questionable? Alternative explanations (confounders via `JOIN`)?
    \item \textbf{Framing Challenges:} Right question for the problem? Neglects perspectives/temporal frames? Aggregation level suitable?
    \item \textbf{Implementation Challenges:} Feasibility issues or unintended consequences suggested by data?
\end{enumerate}

\section{Experimental Setup Details}
\label{sec:experimental-setup}

\subsection{Interactive User Study Questionnaire}
\label{subsec:user-study}
We designed an intuitive questionnaire to assess user experience with VeriMinder across four key dimensions: scenario realism, suggestion effectiveness, rationale clarity, and impact on analysis. Users rated each dimension on a 5-point Likert scale. Figure 9 shows the feedback form used in our interactive study.

\begin{figure}[ht!]
\centering
\fbox{\includegraphics[width=\columnwidth]{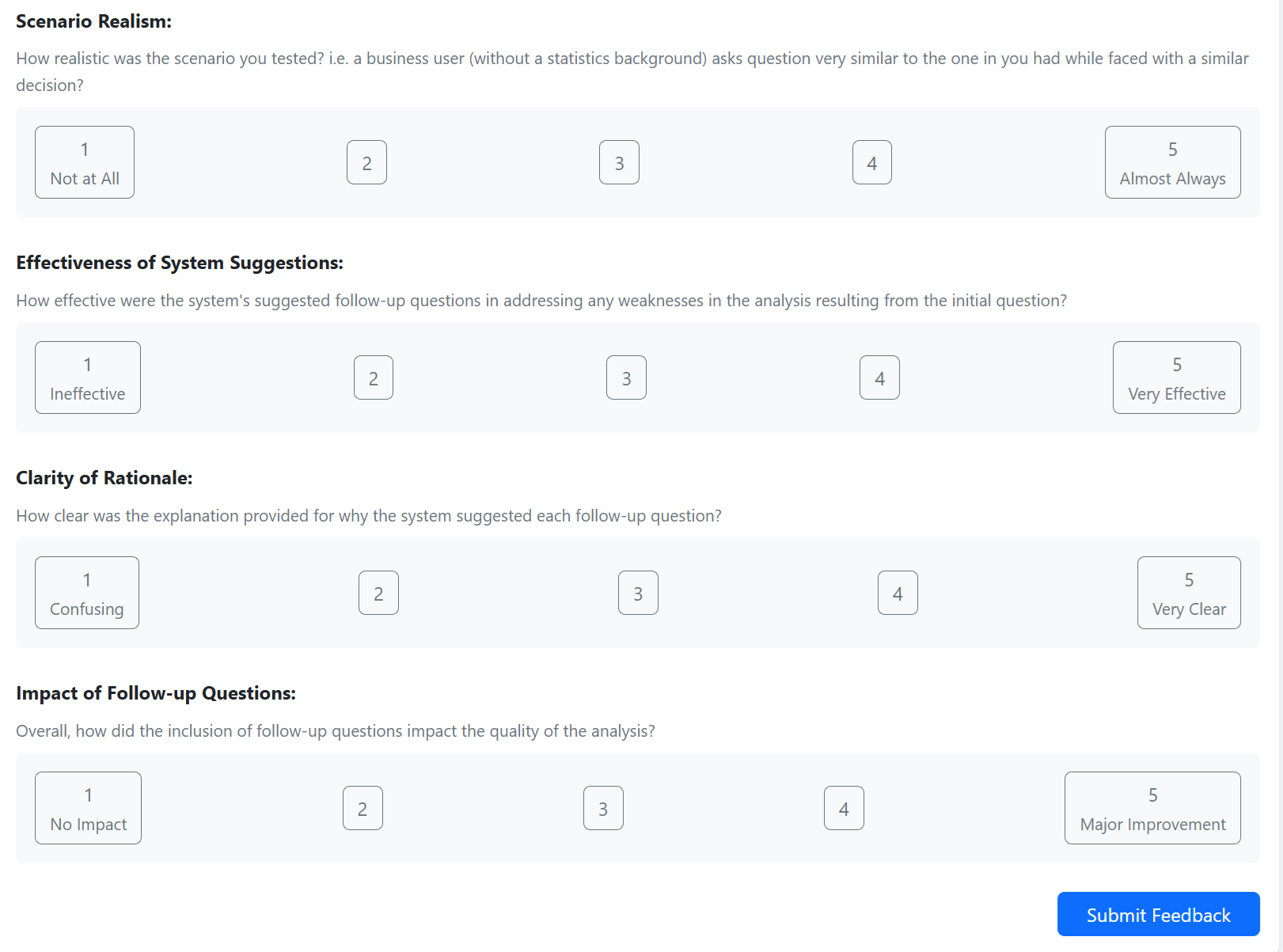}}
\caption{Interactive user study feedback interface}
\label{fig:user_feedback}
\end{figure}
\begin{figure}[ht!]
\centering
\fbox{\includegraphics[width=\columnwidth]{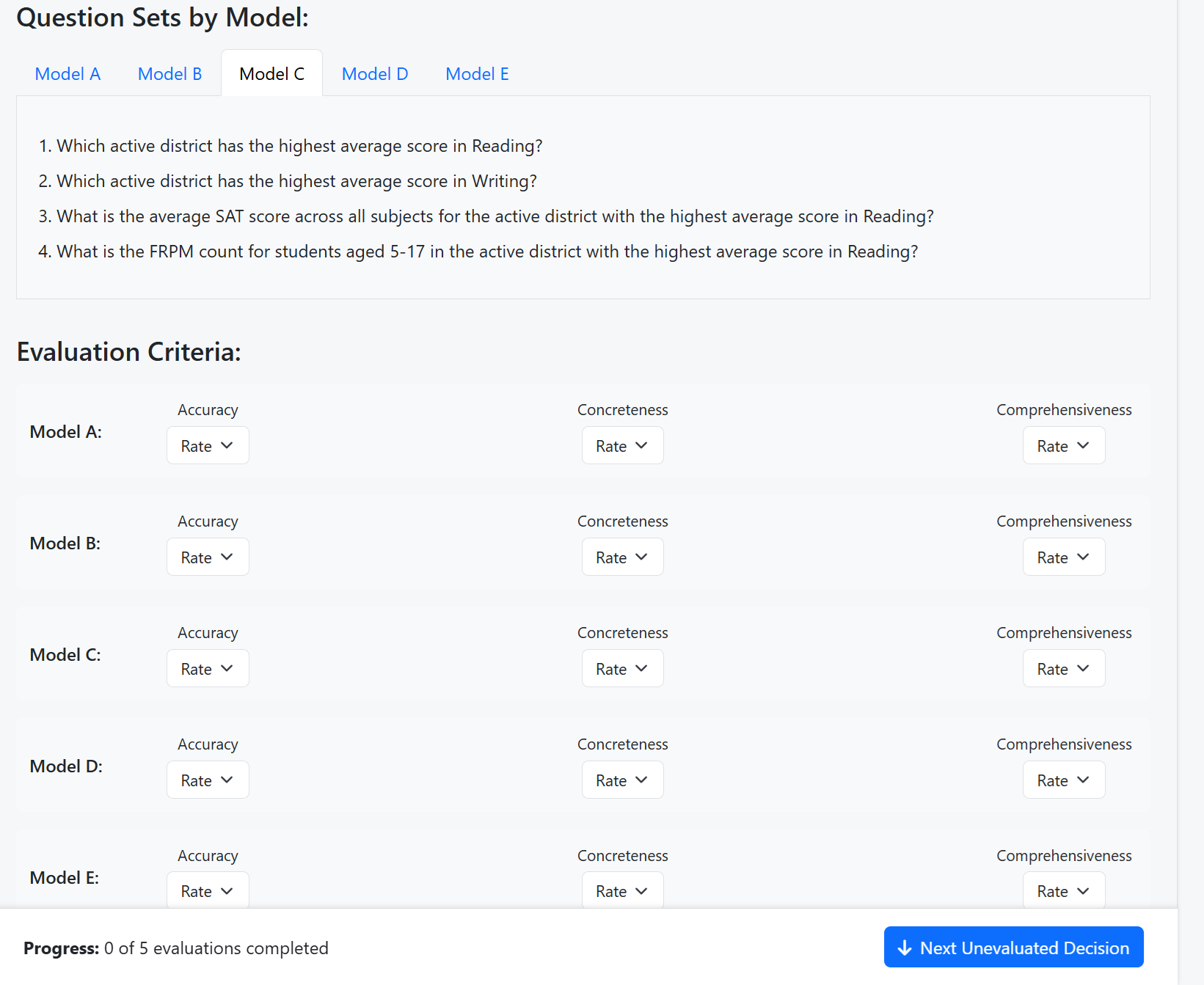}}
\caption{Comparative evaluation interface for assessing analytical quality across methods}
\label{fig:comparative_eval}
\end{figure}

\subsection{Comparative System Evaluation}
\label{subsec:comparative-eval}
The comparative evaluation required participants to rate all five systems (VeriMinder, Direct NL2SQL, Decision-Focused Query Generation, Question Perturbation, and Critic-Agent Feedback - with names anonymized during the testing) on three analytical dimensions: accuracy, concreteness, and comprehensiveness. Participants rated each dimension on a 10-point scale for each system, allowing for direct comparison. Figure 10 shows the evaluation interface.

\subsection{Automated Evaluation Procedure}
\label{subsec:automated-eval}
\begin{enumerate}[itemsep=-3pt]
    \item \textbf{Goal:} To assess the analytical quality of query sets generated by VeriMinder and four baseline systems against the large-scale dataset (100 pairs).
    \item \textbf{Methodology:} Employed an LLM evaluator (Gemini Flash 2.0) ~\cite{gemini2025} using a structured prompt that included:
        \begin{enumerate}[itemsep=-3pt]
            \item The decision context and original NL question.
            \item Database schema snippets and relevant evidence context.
            \item The complete set of successfully executed SQL query results generated by \emph{each} of the five systems (VeriMinder, Direct NL2SQL, Decision-Focused, PerQS, CAF) for the given decision scenario. Our choice of LLM was primarily driven by the response time \cite{artificialanalysis2025} and streaming support dictated by our user interface requirements.
        \end{enumerate} 
    \item \textbf{Evaluation Task:} The LLM was instructed to:
        \begin{enumerate}[itemsep=-3pt]
            \item Holistically evaluate each system's \emph{entire set} of queries and results in the decision context.
            \item Assess each system based on \textbf{Data Accuracy - Fidelity of Fetched Results to NL Question Intent}, \textbf{Comprehensiveness}, \textbf{Concreteness}, and \textbf{Overall Usefulness}  in the context of the decision goal.
            \item Apply the \textbf{SLOW} framework (\textbf{S}ure, \textbf{L}ook, \textbf{O}pposite, \textbf{W}orst) ~\cite{OSullivan2019} to identify uncertainties, missing information, alternative interpretations, and potential problematic conclusions for each system's output and the combined analysis.
        \end{enumerate}
    \item \textbf{Output:} The process yielded structured evaluations for each system and a comparative assessment, including relative rankings across the specified analytical dimensions.
\end{enumerate}
\end{document}